\documentclass[10pt, a4paper, twocolumn]{naverlabseurope} 

\usepackage{multicol}
\usepackage{tikz,tkz-kiviat,pgfplots}
\usepackage{microtype}
\usepackage{graphicx}
\usepackage[font=small]{caption}
\usepackage{subcaption}
\usepackage{floatrow}
\usepackage{hyperref}
\usepackage{url}
\usepackage{booktabs}
\usepackage{multirow}
\usepackage{array}
\usepackage{wrapfig}
\usepackage{makecell}
\usepackage{listings}
\usepackage{inconsolata}

\title{ELITR-Bench: A Meeting Assistant Benchmark for Long-Context Language Models}

\authors{Thibaut Thonet \authsep Jos Rozen \authsep Laurent Besacier}
\affiliations{NAVER LABS Europe}
\contributions{}
\website{}
\websiteref{\texttt{\{thibaut.thonet,jos.rozen,laurent.besacier\}@naverlabs.com}}

\newcolumntype{C}{@{\extracolsep{3cm}}c@{\extracolsep{0pt}}}%
\newfloatcommand{capbtabbox}{table}[][\FBwidth]

\begin{abstract}
Research on Large Language Models (LLMs) has recently witnessed an increasing interest in extending the models' context size to better capture dependencies within long documents. While benchmarks have been proposed to assess long-range abilities, existing efforts primarily considered generic tasks that are not necessarily aligned with real-world applications. In contrast, we propose a new benchmark for long-context LLMs focused on a practical meeting assistant scenario in which the long contexts consist of transcripts obtained by automatic speech recognition, presenting unique challenges for LLMs due to the inherent noisiness and oral nature of such data. Our benchmark, ELITR-Bench, augments the existing ELITR corpus by adding 271 manually crafted questions with their ground-truth answers, as well as noisy versions of meeting transcripts altered to target different Word Error Rate levels. Our experiments with 12 long-context LLMs on ELITR-Bench confirm the progress made across successive generations of both proprietary and open models, and point out their discrepancies in terms of robustness to transcript noise. 
We also provide a thorough analysis of our GPT-4-based evaluation, including insights from a crowdsourcing study. Our findings indicate that while GPT-4's scores align with human judges, its ability to distinguish beyond three score levels may be limited.
\end{abstract}

\begin{document}

\maketitle

\section{Introduction}

The context window of Large Language Models (LLMs) has recently undergone a significant expansion, scaling from a few thousand tokens to tens or even hundreds of thousands~\citep{chen2023extending,xiong2023effective,Liu2023ring,chen2024longlora,longalign}. As a consequence, benchmarks have emerged to assess LLMs' long-range abilities, tackling the specific challenges of Question Answering (QA) on long documents~\citep{LEval,longchat2023,longbench,loogle,maharana2024evaluating,infinitybench}. However, while previous datasets focusing on long-context models offer longitudinal evaluations across different tasks, they often provide only superficial analyses of each task. The covered tasks are also often generic~-- e.g., questions on Wikipedia~\citep{loogle}~-- and thus not particularly suitable for realistic, focused applications.\footnote{WildBench \cite{wildbench} is a very recent exception as it proposes an automated evaluation framework for assessing LLMs with complex real-world user queries.}

In contrast, our work advocates for a situated evaluation of long-context LLM performance within specific, real-world scenarios. As a practical illustration, consider a meeting assistant that allows users to inquire about meetings they did not attend. Given that hour-long meeting transcripts must fit within the agent's context window, proficient handling of long contexts is a prerequisite.
This paper then introduces the first benchmark~-- to the best of our knowledge~-- for evaluating long-context LLMs on a realistic meeting assistant task. Our benchmark, named ELITR-Bench,\footnote{We release the data associated to our benchmark, as well as the generations produced by compared models and their evaluation, at \url{https://github.com/utter-project/ELITR-Bench}.} is built upon the meeting transcripts of the past ELITR project~\citep{nedoluzhko-etal-2022-elitr}. These transcripts have been obtained by Automatic Speech Recognition (ASR) with minimal human corrections~-- yielding long, noisy documents of oral nature that present unique challenges for LLMs. ELITR-Bench enables tracking the progress of successive model generations (e.g., GPT-3.5 vs GPT-4/4o, or LLaMA-2 vs LLaMA-3.1), as demonstrated by our extensive experiments on 12 long-context proprietary and open LLMs. Additionally, we analyzed how these models handle varying noise levels in meeting transcripts. 
Our GPT-4-based evaluation was validated through a crowdsourcing study, which revealed a high correlation with human judges' scores but also a limited ability to distinguish between more than three quality levels despite using a score range of 10 points. 

The remainder of the paper is structured as follows. We provide a review of related literature in Section~\ref{sec:related-work}, before introducing the proposed ELITR-Bench in Section~\ref{sec:elitr-bench}. We then describe our experimental setup and results in Sections~\ref{sec:exp-setup} and~\ref{sec:exp-results}, respectively. Section~\ref{sec:llm-eval-assess} provides an in-depth assessment of our LLM-based evaluation methodology. Finally, Section~\ref{sec:conclusion} concludes the paper and provides some perspectives for future work.

\section{Related work}
\label{sec:related-work}

\paragraph{Long-context LLMs and techniques.}

Numerous techniques have emerged to address the challenge of long-context modeling.\footnote{For a comprehensive collection of resources on this subject, we let the reader refer to \url{https://github.com/Xnhyacinth/Awesome-LLM-Long-Context-Modeling}.} While an exhaustive survey of these methods is beyond the scope of this paper, they can generally be categorized into three main groups (excluding other distinct approaches such as retrieval-augmented generation~\citep{xu2024retrieval} and context compression~\citep{DBLP:conf/emnlp/ChevalierWAC23}): (a) the development of efficient transformer architectures to address the quadratic attention challenge, including sparse transformers \citep{DBLP:journals/corr/abs-1904-10509,DBLP:journals/corr/abs-2004-05150,DBLP:journals/corr/abs-2007-14062,DBLP:journals/corr/abs-2006-07214}, linear transformers \citep{DBLP:journals/corr/abs-2006-16236,DBLP:journals/corr/abs-2006-04768,DBLP:journals/corr/abs-2009-14794}, and hierarchical transformers \citep{DBLP:journals/corr/abs-2201-06774,DBLP:journals/corr/abs-2106-01040,liu2022erniesparse}; (b) approaches like recurrent attention networks \citep{DBLP:journals/corr/abs-1901-02860,peng2023rwkv,bulatov2024scaling} and state-space models \citep{gu2023mamba,wang2024mambabyte}; (c) length extrapolation or position embedding interpolation, where LLMs are fine-tuned or adapted at inference time to adjust tokens' positions to match the new context length \citep{chen2023extending,xiong2023effective,peng2023yarn,pal2023giraffe,Liu2023ring,chen2024longlora,longalign}.
These techniques also contributed to the context length expansion in proprietary models like GPT-4~(32k-128k), Claude-3~(200k), and Gemini-1.5~(128k-1M).

\paragraph{Long-context benchmarks.}

Several benchmarks have recently emerged with the growing interest in evaluating techniques that extend the context length of LLMs. Long Range Arena~\citep{LongRangeArena} was proposed to assess the quality of efficient transformer models in long-context scenarios, covering  1K-16K tokens sequences through different data types and modalities. L-Eval \citep{LEval} offers a comprehensive evaluation suite with 20 sub-tasks and over 2,000 human-labeled query-response pairs, aggregating pre-existing datasets like NarrativeQA~\citep{NarrativeQA}.
LongEval~\citep{longchat2023} proposes synthetic tasks of varying difficulty, while
LongBench~\citep{longbench} and LongBench-Chat~\citep{longalign} aggregate several datasets in English and Chinese.
Other recent benchmarks appeared, such as: LongAlpaca~\citep{chen2024longlora}, Loogle~\citep{loogle}, LoCoMo~\citep{maharana2024evaluating}, BAMBOO~\citep{dong2024bamboo}, FLenQA~\citep{levy2024task}, and $\infty$Bench~\citep{infinitybench} that proposes an average data length over 100K tokens.
Recently, the needle-in-the-haystack test was proposed by \cite{kamradt2023needle}, in which a long-context LLM must retrieve a short text (the needle) from a long document (the haystack). This initial test has since inspired several subsequent works that propose more and more complex tasks.
%
%
Our contribution, ELITR-Bench, distinguishes itself from existing benchmarks in several ways: (a) it focuses on a real use-case~-- meeting assistants,\footnote{MeeQA~\cite{meeqa} and MeetingQA~\cite{meetingqa} study QA on meeting transcripts, but  questions and answers are directly extracted from the transcripts, leading to lower quality and requiring simple extraction rather than inference, making the task less challenging than ELITR-Bench.} (b) it challenges models by requiring them to make inferences from noisy ASR-based documents, and (c) it offers both question answering and conversation versions~(see Section~\ref{sec:elitr-bench}), enabling the analysis of different prompt modes.




\paragraph{Evaluation with LLMs.} 

Recent works explored the use of LLMs such as GPT-4 as judges to evaluate responses on open-ended questions. \citet{zheng2023judging} measured agreement between LLM and human evaluators while introducing two datasets (MT-bench and Chatbot Arena).
They showed that LLM judges like GPT-4 can match both controlled and crowdsourced human annotations, achieving over 80\% agreement~-- the same level of agreement between humans.
\citet{he2024if} evaluated the performance of GPT-4 against 415 crowdsourcing human labelers.
Despite employing best quality control practices, the highest labeling accuracy achieved through crowdsourcing was 81.5\% whereas GPT-4 obtained 83.6\%. 
As in certain scenarios, employing proprietary LLMs as evaluators can pose challenges due to their closed-source nature, \citet{kim2023prometheus} introduced Prometheus, an open-source LLM fine-tuned for evaluation. 
Recently, \citet{bavaresco2024llmsinsteadhumanjudges} introduced Judge-Bench, a collection of 20 NLP datasets with human annotations for evaluating LLMs' ability to replicate human judgments.
In this work, we compare LLMs-as-a-judge~(GPT-4 and Prometheus) with expert and crowdsourcing-based human evaluators to assess responses generated by several long-context models on ELITR-Bench. 



\section{ELITR-Bench}
\label{sec:elitr-bench}

We build our benchmark on top of the ELITR Minuting Corpus \citep{nedoluzhko-etal-2022-elitr}.\footnote{Accessible at: \url{https://lindat.mff.cuni.cz/repository/xmlui/handle/11234/1-4692}} This corpus contains transcripts of meetings conducted in both Czech and English, along with manually crafted summaries referred to as `minutes'. The meeting durations range from 10 minutes to over 2 hours, with the majority lasting around one hour. Although transcripts have been corrected from ASR outputs, they still contain noise and reflect various oral language phenomena such as interjections. Each transcript is de-identified\footnote{The authors ensured the removal or masking of any personally identifiable information (PII), such as names, addresses, or other details from the transcripts. Moreover, they de-identified any project or organization-related information, as its inclusion could indirectly reveal the individuals involved.} and accompanied by one or multiple corresponding minutes files. However, in the benchmark described here, we only use the verbatim transcripts and exclude the minutes. In the current version of ELITR-Bench, our focus is on English meetings. Specifically, we utilized the official \textit{dev} and \textit{test2} sets, consisting of 10 and 8 meetings respectively, both sourced from ELITR-English. These meetings focus on discussions related to the computer science domain, with a particular emphasis on Natural Language Processing (NLP) topics. For every meeting within this corpus, we manually created a series of questions that can be directly addressed using the corresponding transcript, and provided their corresponding ground-truth answers. We present in Appendix~\ref{app:excerpt} (Table~\ref{tab:elitr-example}) a snippet of an ELITR meeting transcript, and showcase examples of questions and answers introduced in ELITR-Bench.

\paragraph{Question type and answer position.} The questions we defined span various types, including:
\textbf{Who} questions,
\textbf{What} questions (which also include \textit{Why} questions),
\textbf{When} questions,
and \textbf{How many} questions.
Additionally, we annotated the position of the answer within the meeting transcript, categorizing it as either in the \textbf{Beginning} (first third), \textbf{Middle} (second third), \textbf{End} (final third), or spanning \textbf{Several} passages throughout the transcript. Table~\ref{tab:elitr-plus} provides a summary of the statistics for our benchmark.

\paragraph{QA and Conversation settings.} The proposed ELITR-Bench is available in two settings. In \textbf{ELITR-Bench-QA}, we designed for each meeting a set of stand-alone questions (along with their answers) that can be addressed solely based on the meeting transcript, without additional context. We also designed a modified \textbf{ELITR-Bench-Conv} version where questions are to be asked in sequence, in a pre-defined order within a conversation. In this setting, some of the questions contain pronominal references or ellipses, for which previous conversational context (i.e., previous questions and answers) must be used to answer properly. For example, the question ``\textit{What is challenging about testing the demo system at the students firm fair?}'' from the QA setting is replaced in the Conv setting with ``\textit{What is challenging about this event?}'', where the answer to the previous question in the conversation was ``\textit{The students firm fair}''. Such questions have been obtained by manually re-writing the Conv questions into QA questions by resolving coreferences. The number of QA/Conv differentiating questions is 16 (out of 141) for the dev set and 17 (out of 130) for the test set.


\begin{table*}[t]
\caption{Statistics for the ELITR-Bench dataset: all questions and answers are annotated by question type (\textit{What}, \textit{Who}, \textit{When}, \textit{How many}) and by the position of the answer within the meeting transcript (\textit{Beginning}, \textit{Middle}, \textit{End}, or spanning \textit{Several} sections). The number of tokens per meeting is counted using a LLaMA-2 tokenizer.}
\label{tab:elitr-plus}
\centering
\scalebox{0.83}{
\begin{tabular}{lrrlrlrr}
\toprule
\textbf{Split} & \multicolumn{1}{l}{\textbf{\#Meetings}} & \multicolumn{1}{l}{\textbf{\#Questions}} & \multicolumn{2}{l}{\textbf{\begin{tabular}[c]{@{}l@{}}\#Questions by\\ question type\end{tabular}}} & \multicolumn{2}{l}{\textbf{\begin{tabular}[c]{@{}l@{}}\#Questions by\\ answer position\end{tabular}}} & \multicolumn{1}{l}{\textbf{\begin{tabular}[c]{@{}l@{}}\#Tokens per meeting:\\ average [min, max]\end{tabular}}} \\ \midrule
Dev & 10 & 141 & What & 59 & Begin & 45 & 11.3k [5.2k, 17.4k] \\
 &  &  & Who & 51 & Middle & 29 & \\
 &  &  & When & 21 & End & 32 & \\
 &  &  & How many & 10 & Several & 35 & \\ \midrule
Test & 8 & 130 & What & 57 & Begin & 43 & 12.6k [4.8k, 17.6k] \\
 &  &  & Who & 45 & Middle & 34 & \\
 &  &  & When & 20 & End & 22 & \\
 &  &  & How many & 8 & Several & 31 & \\ \bottomrule
\end{tabular}
}
\end{table*}

\paragraph{Noisy versions of the meeting transcripts.} To assess the robustness of long-context LLMs to noisy text, we created multiple noisy versions of the ELITR meeting transcripts by simulating different levels of ASR noise. Using a large corpus of over 500k ASR transcripts aligned with references,\footnote{\url{https://huggingface.co/datasets/google/red_ace_asr_error_detection_and_correction}} we generated 86,148 substitution rules where each rule consists of a token and a probability distribution over similar tokens (or an empty character) that can plausibly replace it. This extensive rule set enables us to simulate noisy transcripts with varying target Word Error Rate levels (20\%, 40\%, 60\%, 80\% and 100\%).\footnote{In practice, the effective Word Error Rate obtained after applying the noise injection procedure to the transcripts is generally lower than the target one~-- it is therefore still possible to infer some answers with a target level of 100\%.} We provide more details in Appendix~\ref{app:noise}. 

\section{Experimental setup}
\label{sec:exp-setup}


\paragraph{Evaluation protocol.} The evaluation on ELITR-Bench is conducted as follows. For each meeting, a prompt containing the transcript and detailing the assistant's task is formed. 
Then, questions are appended to the initial prompt to drive the conversation about the corresponding meeting. We consider two ways to do this: (i) the \textit{single-turn mode}, where only a single question is tackled in the conversation~(i.e., the prompt is re-initialized for each new question), or (ii) the \textit{multi-turn mode}, where all the questions related to a meeting are asked successively within a single conversation. Given the stand-alone nature of questions in ELITR-Bench-QA, one can adopt either the single-turn or multi-turn modes for this setting, whereas for ELITR-Bench-Conv it only makes sense to use the multi-turn mode as some questions are inter-dependent. In our evaluation methodology, given a question integrated in the aforementioned prompt, the response generated by an LLM is evaluated automatically using a GPT-4 judge,\footnote{Our GPT-4 judge is based on the gpt-4-0613 checkpoint, for its cheaper cost compared to gpt-4-turbo models. Pilot experiments with different GPT-4 judges led to similar evaluation scores.} following the standard practice in LLM evaluation (as discussed in Section~\ref{sec:related-work}). Specifically, we adopted a score rubric-based evaluation methodology~\citep{kim2023prometheus} in which a generated response is evaluated on its proximity to the ground-truth answer, given the associated question and a score rubric that details the quality criteria expected at each score level (ranging from the lowest score of 1 to the perfect score of 10). The prompt used for the evaluation as well as our manually defined score rubric are provided in Appendix~\ref{app:prompt} (Figs.~\ref{fig:eval-prompt-gpt4} and~\ref{fig:score-rubric-gpt4}, respectively). Although our core experiment results rely on automatic LLM-based evaluation (Section~\ref{sec:exp-results}), we further confirm the validity of this methodology against human judgement in Section~\ref{sec:llm-eval-assess}.

\paragraph{Compared models.} 

In our experiments on ELITR-Bench, we compared responses generated by 12 LLMs with long-context capabilities (at least 32k tokens). We included both commercial models and open long-context models based on LLaMA-\{2, 3.1\} and Phi-3 in our benchmarking:
\begin{itemize}
    \item \textbf{GPT-3.5}, \textbf{GPT-4}~\citep{OpenAI2023}, \textbf{GPT-4o};
    \item LLaMA-2 models extended for long-context scenarios:  \textbf{LongAlpaca-\{7B, 13B\}} \citep{chen2024longlora}, 
    \textbf{LongChat-7B-v1.5} \citep{longchat2023}, \textbf{Vicuna-\{7B, 13B\}-v1.5}~\citep{vicuna2023}, \textbf{LongAlign-\{7B, 13B\}} \citep{longalign};
    \item \textbf{LLaMA-3.1-8B}~\citep{llama3};
    \item \textbf{Phi-3-small}~\citep{phi3}.
\end{itemize}
We provide more details on the compared models in Appendix~\ref{app:compared-models}. Additionally, we describe the search conducted to select the best model configuration (including inference hyperparameters and prompt formatting) in Appendix~\ref{app:hyperparam}.

\section{Experimental results}
\label{sec:exp-results}

This section describes the results of the experiments conducted on ELITR-Bench. In Section~\ref{sec:main-results}, we summarize the benchmarking results of the compared models on ELITR-Bench-QA and ELITR-Bench-Conv. Then, in Section~\ref{sec:qtype-aposition}, we analyze the impact of the question types and answer positions on the models' performance.
Finally, Section~\ref{sec:noise} discusses how the noise level of the meeting transcripts influences the performance of the models.

\subsection{Main results}
\label{sec:main-results}

The main results of the benchmarking on ELITR-Bench are reported in Table~\ref{tab:main-results}. The compared models are evaluated in three settings that combine the ELITR-Bench-QA or ELITR-Bench-Conv question set with the single-turn mode (i.e., one question asked per conversation) or multi-turn mode (i.e., all questions related to one meeting asked in a single conversation).\footnote{Single-turn ELITR-Bench-Conv is omitted as some questions in the Conv setting are context-dependent (i.e., rely on previous questions or answers) and thus could not be asked independently.} For each of the three considered settings, we report the results on the dev set, the results on the test set, and their mean. Given the extensive cost of GPT4-based evaluation (detailed in Section~\ref{sec:exp-setup}), we performed a single seeded run for the dev set and three seeded runs for the test set. For the latter, we report the average score over the three runs. In Appendix~\ref{app:test-variance}, we provide more details about the seeded runs as well as their standard deviations.

Looking at the three settings in Table~\ref{tab:main-results}, we observe that GPT-4 and GPT-4o dominate over all other approaches with an average score that is always above 8.\footnote{While one might argue that GPT-4 is advantaged due to the use of a GPT-4 judge, we show in Section~\ref{sec:evaluator-results} that this model's dominance is observed for other evaluators as well.} GPT-3.5 obtained a slightly lower average score~-- around 7~-- and was also beaten by the two more recent open LLMs, LLaMA-3.1-8B and Phi-3-small with LLaMA-3.1-8B coming out on top. These three models notably outperformed the LLaMA-2-based LLMs.
Among the latter, differences are smaller with scores close to 6 on the single-turn setting, and ranging from 4 to 6 on the multi-turn settings. Nonetheless, we can note that Vicuna-13B-v1.5 is the LLaMA-2-based approach that performed the most favorably overall on the three settings. Interestingly, the results in the single-turn and multi-turn modes show large discrepancies for LLaMA-2-based models~-- even when the question set is exactly the same, for ELITR-Bench-QA. This seems to indicate that these LLMs get distracted by the previous questions and answers, which affects their performance. In contrast, GPT-4/4o is able to maintain its performance between the single-turn mode and the multi-turn mode. The same can be observed for LLaMA-3.1-8B and Phi-3-small, suggesting that recent open long-context LLMs are able to successfully handle multi-turn conversations, unlike their predecessors.
Comparing the results of the QA and Conv settings in the multi-turn mode, we found only minimal differences. This can be explained by the small number of questions that differ between QA and Conv (16 for the dev set and 17 for the test set). In Appendix~\ref{sec:differentiating-questions}~(Fig.~\ref{fig:differentiating-questions}), we further analyze the impact of the benchmark setting (QA vs Conv) and inference mode (single-turn vs multi-turn) by detailing the results restricted to this subset of differentiating questions.

\begin{table*}[t]
\centering
\caption{Results on different ELITR-Bench settings. The reported numbers correspond to the average scores from 1 to 10 (higher is better) obtained by a GPT-4 evaluator, on a single seeded run for the dev set and 3 seeded runs for the test set. Boldface numbers correspond to the best performance among proprietary or open models. The results for GPT-3.5 are omitted in the multi-turn setting as the context length exceeded the 16k limit of this model.}
\label{tab:main-results}
\scalebox{0.87}{
\begin{tabular}{lrrrrrrrrr}
\toprule
\multirow{3}{*}[-0.15cm]{\textbf{Model}} & \multicolumn{3}{c}{\textbf{Single-turn}} & \multicolumn{6}{c}{\textbf{Multi-turn}} \\ \cmidrule(l{4pt}r{4pt}){2-4} \cmidrule(l{4pt}r{4pt}){5-10} 
 & \multicolumn{3}{c}{\makebox[3cm]{\textbf{ELITR-Bench-QA}}} & \multicolumn{3}{c}{\makebox[3cm]{\textbf{ELITR-Bench-QA}}} & \multicolumn{3}{c}{\makebox[3cm]{\textbf{ELITR-Bench-Conv}}} \\ \cmidrule(l{4pt}r{4pt}){2-4} \cmidrule(l{4pt}r{4pt}){5-7} \cmidrule(l{4pt}r{4pt}){8-10} 
 & \textbf{Dev} & \textbf{Test} & \textbf{Mean} & \textbf{Dev} & \textbf{Test} & \textbf{Mean} & \textbf{Dev} & \textbf{Test} & \textbf{Mean} \\ \midrule
GPT-3.5 & 7.04 & 7.44 & 7.24 & - & - & - & - & - & - \\
GPT-4 & 8.21 & 8.39 & 8.30 & \textbf{8.53} & \textbf{8.42} & \textbf{8.47} & \textbf{8.53} & 8.36 & 8.44 \\
GPT-4o & \textbf{8.53} & \textbf{8.44} & \textbf{8.48} & 8.33 & 8.38 & 8.36 & 8.48 & \textbf{8.41} & \textbf{8.45} \\ \midrule
LongAlpaca-7B & 5.89 & 5.60 & 5.75 & 4.53 & 4.84 & 4.68 & 4.70 & 4.58 & 4.64 \\
LongAlpaca-13B & 6.17 & 6.25 & 6.21 & 4.76 & 4.71 & 4.73 & 4.74 & 4.74 & 4.74 \\
LongChat-7B-v1.5 & 6.60 & 5.78 & 6.19 & 5.85 & 4.17 & 5.01 & 5.21 & 4.31 & 4.76 \\
Vicuna-7B-v1.5 & 5.42 & 5.61 & 5.51 & 4.68 & 4.61 & 4.65 & 4.67 & 4.69 & 4.68 \\
Vicuna-13B-v1.5 & 5.92 & 6.52 & 6.22 & 5.52 & 5.67 & 5.60 & 5.42 & 5.78 & 5.60 \\
LongAlign-7B & 6.11 & 6.46 & 6.28 & 5.43 & 4.47 & 4.95 & 5.04 & 5.06 & 5.05 \\
LongAlign-13B & 6.27 & 6.33 & 6.30 & 4.65 & 5.33 & 4.99 & 4.81 & 4.95 & 4.88 \\
LLaMA-3.1-8B & \textbf{7.70} & \textbf{7.83} & \textbf{7.76} & \textbf{7.77} & \textbf{7.81} & \textbf{7.79} & \textbf{7.80} & \textbf{7.78} & \textbf{7.79} \\
Phi-3-small & 7.31 & 7.34 & 7.32 & 7.67 & 7.52 & 7.59 & 7.53 & 7.38 & 7.46 \\ \bottomrule
\end{tabular}
}
\end{table*}

\subsection{Impact of question type and answer position} 
\label{sec:qtype-aposition}

\begin{table*}[t]
\centering
\caption{Results by question type and answer position on the test set of ELITR-Bench-QA in single-turn mode. The reported numbers correspond to the average scores from 1 to 10 (higher is better) obtained by a GPT-4 evaluator. The number N below a subset indicates the corresponding subset size. Boldface numbers correspond to the best performance among proprietary or open models.}
\label{tab:qtype-aposition-full}
\scalebox{0.87}{
\begin{tabular}{lrrrrrrrr}
\toprule
\multirow{2}{*}[-0.2cm]{\textbf{Model}} & \multicolumn{4}{c}{\textbf{Question type}} & \multicolumn{4}{c}{\textbf{Answer position}} \\ \cmidrule(l{4pt}r{4pt}){2-5} \cmidrule(l{4pt}r{4pt}){6-9}
 & \textbf{\begin{tabular}[c]{@{}r@{}}Who\\ (N=45)\end{tabular}} & \textbf{\begin{tabular}[c]{@{}r@{}}What\\ (N=57)\end{tabular}} & \textbf{\begin{tabular}[c]{@{}r@{}}When\\ (N=20)\end{tabular}} & \textbf{\begin{tabular}[c]{@{}r@{}}How many\\ (N=8)\end{tabular}} & \textbf{\begin{tabular}[c]{@{}r@{}}Begin\\ (N=43)\end{tabular}} & \textbf{\begin{tabular}[c]{@{}r@{}}Middle\\ (N=34)\end{tabular}} & \textbf{\begin{tabular}[c]{@{}r@{}}End\\ (N=22)\end{tabular}} & \textbf{\begin{tabular}[c]{@{}r@{}}Several\\ (N=31)\end{tabular}} \\ \midrule
GPT-3.5 & 7.91 & 6.94 & 7.68 & 7.79 & 7.33 & 7.45 & 7.76 & 7.37 \\
GPT-4 & 8.56 & \textbf{8.29} & 8.28 & 8.29 & \textbf{8.36} & 8.29 & 8.32 & \textbf{8.57} \\
GPT-4o & \textbf{8.68} & 8.12 & \textbf{8.60} & \textbf{8.92} & 8.17 & \textbf{8.67} & \textbf{8.42} & 8.56 \\ \midrule
LongAlpaca-7B & 5.35 & 5.37 & 6.35 & 6.79 & 5.81 & 5.80 & 4.97 & 5.53 \\
LongAlpaca-13B & 7.19 & 5.47 & 6.47 & 6.00 & 5.93 & 5.95 & 6.85 & 6.59 \\
LongChat-7B-v1.5 & 6.88 & 4.94 & 6.33 & 4.17 & 6.41 & 4.91 & 5.89 & 5.77 \\
Vicuna-7B-v1.5 & 6.13 & 5.65 & 5.40 & 2.88 & 5.89 & 5.21 & 4.96 & 6.12 \\
Vicuna-13B-v1.5 & 6.96 & 6.68 & 5.48 & 5.54 & 6.35 & 6.41 & 6.55 & 6.87 \\
LongAlign-7B & 6.93 & 6.33 & 6.00 & 5.88 & 7.09 & 6.39 & 6.47 & 5.66 \\
LongAlign-13B & 6.08 & 6.74 & 5.97 & 5.75 & 6.71 & 6.21 & 6.33 & 5.95 \\
LLaMA-3.1-8B & \textbf{8.18} & \textbf{7.53} & 7.53 & \textbf{8.67} & \textbf{7.95} & \textbf{7.60} & \textbf{8.00} & \textbf{7.77} \\
Phi-3-small & 7.67 & 6.78 & \textbf{7.85} & 8.25 & 7.57 & 7.36 & 7.06 & 7.22 \\ \bottomrule
\end{tabular}
}
\end{table*}

In this section, we provide the full results split by question type and answer position obtained on ELITR-Bench-QA's test set in the single-turn setting. The results are reported in Table~\ref{tab:qtype-aposition-full}. Looking at the global model performance over the different question types and answer positions, we do not identify any clear trend highlighting a question type or position answer as notably easier or harder. However, the \textit{Who} questions seem to be on average slightly easier to answer. In contrast, the \textit{What} questions were comparatively more challenging than other types for the best performing models (GPT-3.5, GPT-4, GPT-4o, LLaMA-3.1-8B, and Phi-3-small). This is not surprising as \textit{What} questions sometimes require complex answers that go beyond simply listing entities, dates or numbers. Interestingly, LLaMA-2-based models struggled the most with the \textit{How many} questions. Although the amount of such questions is very limited (8 in the test set) which calls for caution on tentative interpretations, this seems to suggest that LLaMA-2 models are notably less proficient at dealing with quantities and numbers than GPT models and more recent open LLMs such as LLaMA-3.1 and Phi-3.

Regarding the answer position, we did not notice any strong ``lost in the middle'' effect \citep{liu2023lost} which posits that information located in the middle section of long contexts is harder to access for LLMs. We verified this by conducting a statistical significance test on the scores obtained by each individual model, whose details are described in Appendix~\ref{app:lost-in-the-middle}.

\subsection{Robustness to transcript noise}
\label{sec:noise}

\begin{figure}[t]
\centering
\includegraphics[width = 0.97\textwidth]{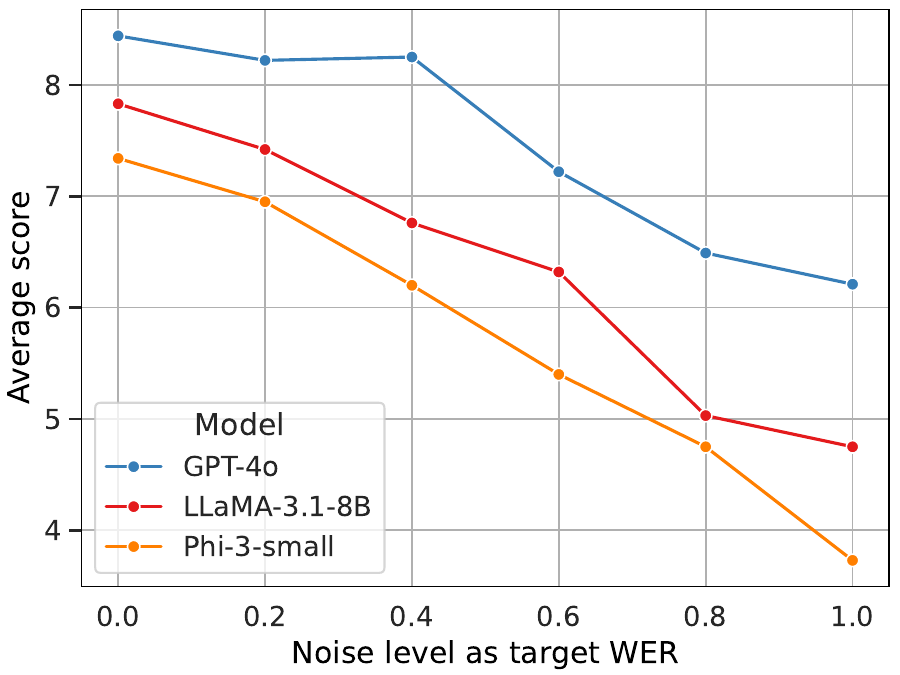} 
\caption{Comparison of the scores obtained for GPT-4o, LLaMA-3.1-8B, and Phi-3-small using transcripts with varied levels of noise on the test set of ELITR-Bench-QA in single-turn mode. Indicated levels of noise correspond to the target Word Error Rates set in our noise injection procedure.}
\label{fig:noise}       
\end{figure}

In Section~\ref{sec:main-results}, we studied how long-context LLMs fare at answering questions when having access to relatively clean meeting transcripts. However, in many practical scenarios, the quality of the transcripts might be degraded due to different factors: e.g., the audio recording conditions, the presence of accented speech, or simply the lacking capabilities of the ASR model. We then sought to understand how robust long-context LLMs are in the presence of a noisy transcript. For that purpose, we tested the three best-performing models from Table~\ref{tab:main-results}~-- i.e, GPT-4o,\footnote{Given the similar performance of GPT-4 and GPT-4o, we only retained GPT-4o due to its lower cost.} LLaMA-3.1-8B, and Phi-3-small~-- on the test set of ELITR-Bench-QA in single-turn mode, using transcripts with varied levels of noise. The noisy transcripts were obtained following the procedure introduced in Section~\ref{sec:elitr-bench} and detailed in Appendix~\ref{app:noise}. The results are reported in Fig.~\ref{fig:noise}. In this experiment, a single seed is used to limit the cost incurred by GPT-4-based evaluation.

We observe in Fig.~\ref{fig:noise} that while the gap between the two open models and GPT-4o is small on the clean transcript (around 1 point), it widens significantly as the noise level is increased. Interestingly, GPT-4o also seems to resists much better to mild noise (0.2 and 0.4) in comparison to other models. Even at very high noise levels (0.8 and 1.0), its average score remains above 6 which is similar to the performance of LLaMA-2-based models from Table~\ref{tab:main-results}. 
All in all, we conclude that while the most recent open long-context LLMs approach GPT-4/4o capabilities on clean transcripts, there remains an important gap when noisier context is used. 

\section{LLM-based evaluation assessment}
\label{sec:llm-eval-assess}

In this section, we seek to verify the validity of the LLM-based (namely, GPT-4-based) evaluation methodology introduced in Section~\ref{sec:exp-setup} and applied in Section~\ref{sec:exp-results}. In Section~\ref{sec:evaluators}, we define the LLM-based and human-based evaluators that we considered for comparison. The details of the crowdsourcing study we conducted to collect human score annotations are provided in Appendix \ref{app:prolific}. Then, Section~\ref{sec:evaluator-results} presents our results and findings on the evaluator comparison.

\subsection{Compared evaluators}
\label{sec:evaluators}

Our evaluation assessment experiment consists in checking the validity of the numeric scores (from 1 to 10) assigned for each tuple composed of a question, its ground-truth answer, and an LLM response to evaluate. For that purpose, we compared the score annotations obtained through two LLM-based evaluators and two human-based evaluators:
\begin{itemize}
    \item \textbf{GPT-4}~\citep{OpenAI2023}: This evaluator corresponds to the one detailed in the evaluation protocol in Section~\ref{sec:exp-setup} and is based on the gpt-4-0613 model.
    \item \textbf{Prometheus}~\citep{kim2023prometheus}: This fine-tuned model was originally proposed to provide an open-source alternative to using GPT-4 for score rubric-based evaluation. We used the Prometheus-13B-v1.0\footnote{\url{https://huggingface.co/kaist-ai/prometheus-13b-v1.0}} model, with a prompt similar to the one adopted for GPT-4~-- the only difference is that the score rubric is re-scaled to a 1-5 range to fit Prometheus' expected format and multiplied by 2 in post-processing to be comparable to other scores. The prompt and the score rubric are detailed in Appendix \ref{app:prompt} (Figs.~\ref{fig:eval-prompt-prometheus} and~\ref{fig:score-rubric-prometheus}, respectively).
    \item \textbf{Gold Human}: This expert human annotation was done by one of the authors. The scores were assigned following the same 10-point score rubric as the one used for the GPT-4 evaluator (given in Appendix~\ref{app:prompt}, Fig.~\ref{fig:score-rubric-gpt4}), to enforce consistency across questions.
    \item \textbf{Silver Human}: This evaluator is based on a crowdsourcing study with the Prolific\footnote{\url{https://www.prolific.com/}} platform where we averaged the scores assigned by 10 human annotators for each question. The annotators were provided with the same score rubric as for GPT-4 and Gold Human. We give more details on this evaluation in Appendix \ref{app:prolific}.
\end{itemize}
Given the costly nature of human annotations, we performed our evaluation assessment on a small subset of the experiments described in Section~\ref{sec:main-results}. We specifically focused on the results of ELITR-Bench-QA's test set in the single-turn mode. We looked at the results of 3 models that performed diversely in this setting: GPT-4, Vicuna-13B-v1.5, and LongAlpaca-7B.\footnote{The crowdsourcing study was conducted in February 2024~-- i.e., before the release of Phi-3 and LLaMA-3.1~-- which is why these more recent models were not included in the evaluation assessment experiments.}

\subsection{Evaluator comparison results}
\label{sec:evaluator-results}

\begin{table*}[t]
\centering
\caption{Comparison of the scores obtained by different evaluators for the responses generated by GPT-4, Vicuna-13B-v1.5, or LongAlpaca-7B. The reported numbers correspond to the average scores from 1 to 10 (higher is better) obtained on ELITR-Bench-QA's test set in the single-turn mode, and for a single seeded run.}
\label{tab:model-level}
\scalebox{0.89}{
\begin{tabular}{lrrrr}
\toprule
\multirow{2}{*}[-0.1cm]{\textbf{Model}} & \multicolumn{4}{c}{\textbf{Evaluator}} \\ \cmidrule(l){2-5} 
 & \textbf{GPT-4} & \textbf{Prometheus} & \textbf{Gold Human} & \textbf{Silver Human} \\ \midrule
GPT-4 & 8.33 & 5.68 & 7.93 & 7.21 \\
Vicuna-13B-v1.5 & 6.69 & 4.80 & 6.19 & 5.80 \\ 
LongAlpaca-7B & 5.57 & 4.46 & 4.55 & 4.72 \\
\bottomrule
\end{tabular}
}
\end{table*}


\begin{figure*}[t]
\centering
    \begin{subfigure}{.57\textwidth}
    \includegraphics[width = \textwidth]{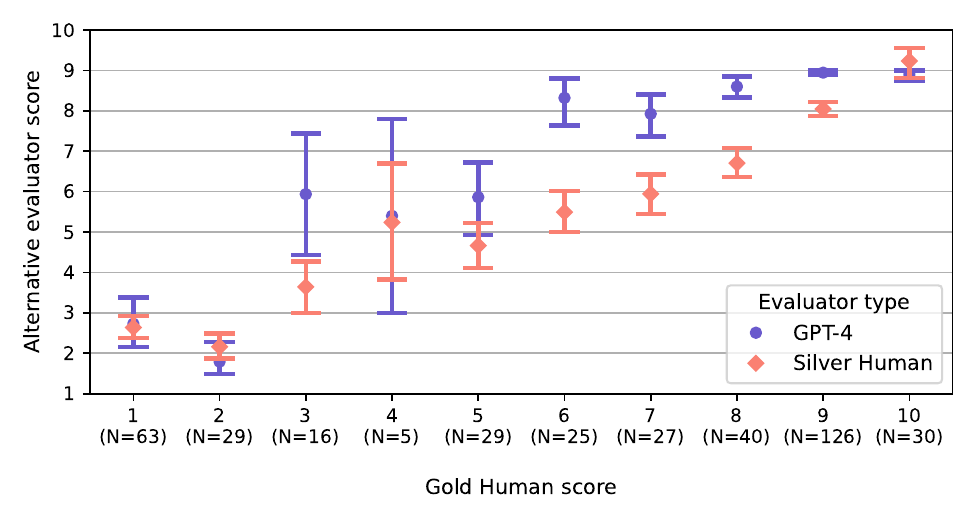} 
    \caption{}
    \label{fig:score-distribution}
    \end{subfigure}
    \hfill
    \begin{subfigure}{.41\textwidth}
    \includegraphics[width = \textwidth]{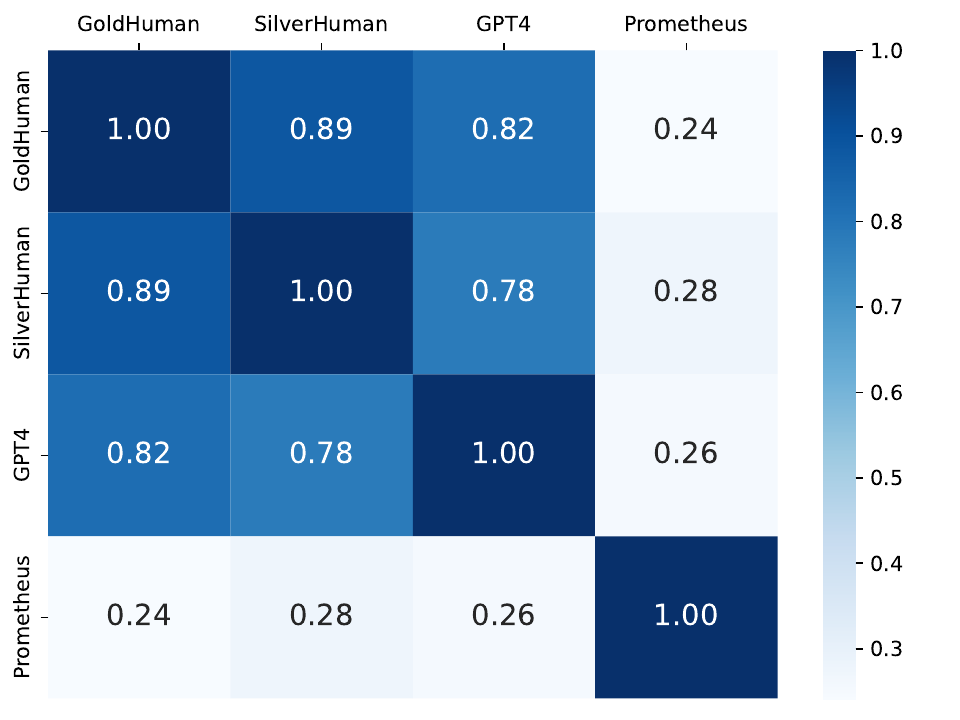} 
    \caption{}
    \label{fig:heatmap}
    \end{subfigure}
\caption{\textbf{(a)} Distribution of GPT-4 and Silver Human scores with respect to each Gold Human score bin (1-10); the N below a score bin indicates the bin size. \textbf{(b)} Pearson correlation between evaluators.}
\label{fig:eval-assessment}       
\end{figure*}

\paragraph{Model-level comparison.} To get a high-level, coarse-grained comparison of the different evaluators introduced above, we applied each of them to the responses generated by GPT-4, Vicuna-13B-v1.5, and LongAlpaca-7B. The results of the corresponding evaluations are presented in Table~\ref{tab:model-level}. We can first observe that the ranking of the three models to evaluate is the same for all the evaluators: GPT-4, Vicuna-13B-v1.5, and LongAlpaca-7B (from the most highly rated model to the most poorly rated one). However, we found that the range of scores was more diverse: Prometheus' scores were overall fairly low (from 4 to 6), while GPT-4's scores are much higher (from 5 to 9). In comparison, the human scores from the Gold Human and Silver Human evaluators were more similar to GPT-4 with scores between 4 and 8.


\paragraph{Correlation analysis.} To get a deeper understanding of how evaluators compare to one another, we calculated the Pearson correlation for every evaluator pair on the responses aggregated over the 8 meetings of the test set and generated by the three retained models. The results are displayed in Fig.~\ref{fig:heatmap}. GPT-4 shows a strong correlation with the two human-based evaluators (0.82 with Gold Human and 0.78 with Silver Human), which is in agreement with the findings from previous studies on GPT-4 judges~\citep{kim2023prometheus,longalign}. Prometheus, on the other hand, yielded a weak correlation~(between 0.2 and 0.3) with all the other evaluators. We hypothesize that this could be due to a domain shift with respect to what Prometheus was fine-tuned on, caused by the nature of the meeting-related questions and the presence of anonymized entities (e.g., [PERSON3]). Turning to the two human-based evaluators, Gold Human and Silver Human obtained a very strong correlation of 0.89 which confirms the validity of the crowdsourcing study and the feasibility of the annotation task by non-expert judges.

\paragraph{Comparison of score distribution across evaluators.} So far in this section, we have found that GPT-4 and human-based evaluators lead to scores that are highly correlated (see Fig.~\ref{fig:heatmap}) but with slightly different score ranges (see Table~\ref{tab:model-level}). This led us to investigate how scores are distributed for different evaluators, and to study to what extent score levels match across evaluators. For that purpose, we considered the pool of (question, response, score) tuples obtained with the Gold Human evaluator on the responses from GPT-4, Vicuna-13B-v1.5, and LongAlpaca-7B for the 130 questions of the test set, i.e., 390 instances in total. We split these instances into 10 bins based on their score value from 1 to 10. Then, for all the instances in a bin, we check the distribution of the scores obtained by other evaluators on the bin's (question, response) pairs. In practical terms, we seek to highlight through this procedure how Gold Human and alternative evaluators align at the grade level. The results are plotted in Fig.~\ref{fig:score-distribution} where we describe the score distribution of the alternative evaluator through its means and 95\% confidence intervals. Interestingly, we observe that the scores for the GPT-4 evaluator seem to fall into 3 distinct clusters, corresponding respectively to the intervals [1, 2], [3, 5] and [6, 10] in the Gold Human scores. This suggests that despite the use of a 10-point score rubric to align the GPT-4 evaluator's scores with detailed desiderata, this evaluator is only able to distinguish between three levels of response quality. This finding then leads us to question the common practice of using LLM-based evaluator scores on a 5-point or 10-point scale. In contrast, the scores from Silver Human show a more linear relationship with the Gold Human scores, suggesting that implementing the 10-point score rubric in the crowdsourcing study aided  in achieving a closer alignment between external human annotators and the evaluation criteria set by the organizers.

\section{Conclusion}
\label{sec:conclusion}

This paper introduced ELITR-Bench, a new benchmark for long-context LLMs focused on the meeting assistant task. We augmented the meeting transcripts from the existing ELITR corpus with 271 manually crafted questions and their respective ground-truth answers. We also produced noisy versions of the transcripts to study the impact of noise on models' performance. Our experiments confirmed the improvements achieved through successive model iterations (e.g., GPT-3.5 vs GPT-4/4o, or LLaMA-2 vs LLaMA-3.1). We also found that while the best tested open models Phi-3 and LLaMA-3.1 approach the performance of OpenAI's frontier model GPT-4o on clean transcripts, there remains a gap in noise robustness between these models.
We validated our evaluation methodology based on a GPT-4 judge through its comparison against a Prometheus-based evaluator, as well as an expert human evaluator and a crowdsourcing-based evaluator. We demonstrate that the GPT-4 judge displays good correlation with human judgments, but a deeper investigation also reveals that it is unable to provide a very fine-grained evaluation on a 10-point scale, contrary to non-expert humans recruited on a crowdsourcing platform.

As future work, we are considering extending ELITR-Bench for the evaluation of retrieval augmented generation (RAG) models. For instance, we could split each transcript into a set of short passages (containing a few utterances) and annotate the relevant passage(s) for each answer. Then, RAG models would generate the response to a question using the retrieved passage(s).
Further studying the impact of de-identification (e.g., named entity anonymization) on QA performance is another interesting direction that could be investigated by reintroducing fake, randomly generated names into the meeting transcripts. We expect that this might have an impact on \textit{Who} questions, which heavily depend on correctly generating anonymized entities that often have an important character overlap (e.g., [PERSON1] and [PERSON11]).

\section*{Acknowledgments}

This paper was partially funded by the European Commission through the UTTER project under grant number 101070631.

\section*{Ethics statement}

The data collection and evaluation process rigorously adhered to the guidelines established by the UTTER EU project. In accordance with EU project policies, we regularly report to an ethics panel, with the most recent Ethical Review meeting held on September 5th, 2024.

Notably, for the human evaluation of LLMs, we chose Prolific, a crowdsourcing platform tailored for academic research. We meticulously followed Prolific's guidelines for human experiments, deviating only in terms of compensation for human labelers. While Prolific sets a minimum compensation of \$6.50 per hour, we offered a significantly higher rate of £9 per hour (equivalent to \$11.5 per hour).

\section*{Reproducibility statement}

The complete set of ELITR-Bench (question, ground-truth answer) pairs, along with metadata, is publicly available at \url{https://github.com/utter-project/ELITR-Bench}. We also indicate for each question the responses generated by the different long-context LLMs considered in this paper, as well as the evaluation score attributed by the GPT-4 judge and other evaluators studied.

Additionally, the code for the response generation and for the evaluation was released alongside the data to enable reproducibility of the paper's results and to foster future benchmarking efforts on ELITR-Bench.

{
    \small
    \bibliographystyle{ieeenat_fullname}
    \bibliography{main}
}

\clearpage

\appendix

\section{Noise injection in meeting transcripts}
\label{app:noise}

To evaluate the robustness of long-context models to noisy text, we generated multiple noisy versions of the ELITR meeting transcripts by simulating various levels of automatic speech recognition (ASR) noise. We utilized a large corpus of over 500,000 ASR transcripts aligned with reference texts, derived from the LibriSpeech corpus \cite{librispeech} and decoded using the Google Cloud Speech-to-Text API. The dataset includes annotated transcription errors and is available from the RED-ACE ASR Error Detection and Correction dataset.\footnote{\url{https://huggingface.co/datasets/google/red_ace_asr_error_detection_and_correction}} We combined the train, development, and test sets into a single JSON file with 525,308 entries, using it to generate 86,148 substitution rules by adapting code from an ASR error simulator.\footnote{\url{https://github.com/pe-trik/asr-errors-simulator}} Each rule specifies a token and a probability distribution over similar tokens (or an empty character) that can replace it. An example of a such a substitution rule is as follows:


\begin{lstlisting}
{
    "earlier": {
        "early": 0.4651, "Elliot": 0.0233, "Early": 0.0233,
        "Hurley": 0.0233, "earliest": 0.1860, "other": 0.0465,
        "earrings": 0.0233, "early.": 0.0233, "area": 0.0233,
        "earlier.": 0.0698, "Julia": 0.0233, "audio": 0.0233,
        "Aaliyah": 0.0233, "Italia": 0.0233
    }
}
\end{lstlisting}

These rules enable the simulation of noisy transcripts from existing `clean' transcripts, targeting specific Word Error Rate (WER) levels of 20\%, 40\%, 60\%, 80\%, and 100\%. However, the actual WER achieved after noise injection is generally lower than the intended target, meaning that even at a target WER of 100\%, some correct answers can still be inferred. The actual WER values corresponding to different target levels are presented in Table \ref{tab:wer}.

\begin{table}[t]
\centering
\caption{Comparison of target Word Error Rates with the effective Word Error Rates obtained by the noise injection procedure, averaged over all dev and test meeting transcripts.}
\label{tab:wer}
\begin{tabular}{rr}
\toprule
\textbf{Target WER (\%)} & \textbf{Effective WER (\%)} \\ \midrule
20.00 & 20.99 \\
40.00 & 37.41 \\
60.00 & 50.07 \\
80.00 & 61.86 \\
100.00 & 71.13 \\ \bottomrule
\end{tabular}
\end{table}

Below is an example of GPT-4's response to the same question when provided with transcripts from the same meeting but with varying levels of noise in the long-context window. The answers below correspond to the question \textit{``Who is going to register for the [PROJECT2] conference?''}:
\begin{itemize}
    \item \textbf{Ground truth:} \textit{``[PERSON3]''}
    \item \textbf{20\% WER:} \textit{``(PERSON3) is going to register for the [PROJECT2] conference.''}
    \item \textbf{40\% WER:} \textit{``The transcript does not explicitly state who will register for the [PROJECT2] conference.''}
    \item \textbf{60\% WER:} \textit{``Based on the transcript, (PERSON3) indicated that they can register for the [PROJECT2] conference.''}
    \item \textbf{80\% WER:} \textit{``Based on the transcript, it is not entirely clear who specifically is going to register for the [PROJECT2] conference.''}
\end{itemize}
We observe that GPT-4 remains robust to text noise across several noise levels for this query. It is important to note that a poor response at 40\% WER and a better response at higher noise levels can occur because the target WERs (noise levels) are calculated for the entire meeting transcript. Various segments of the transcript, especially the relevant ones, may be impacted in diverse ways or remain unaffected at different noise levels.

\section{Experimental setup details}

\subsection{Compared models and hardware details}
\label{app:compared-models}

\begin{table*}[t]
\centering
\caption{Summary of the long-context models compared in Section~\ref{sec:exp-results}. *Vicuna models are provided with a 16k context limit, but it was extended to 32k using RoPE extrapolation~\citep{Su2024rope}.}
\label{tab:compared-models}
\scalebox{0.75}{
\begin{tabular}{lrll}
\toprule
\textbf{Model} & \textbf{\begin{tabular}[c]{@{}r@{}}Context\\ limit\end{tabular}} & \textbf{Backbone} & \textbf{Link} \\ \midrule
GPT-3.5 (turbo-16k-0613) & 16k & - & \url{https://platform.openai.com/docs/models/gpt-3-5-turbo} \\
GPT-4 (1106-preview) & 128k & - & \url{https://platform.openai.com/docs/models/gpt-4-and-gpt-4-turbo} \\
GPT-4o (2024-05-13) & 128k & - & \url{https://platform.openai.com/docs/models/gpt-4-o} \\
LongAlpaca-7B & 32k & LLaMA-2-7B & \url{https://huggingface.co/Yukang/LongAlpaca-7B} \\
LongAlpaca-13B & 32k & LLaMA-2-13B & \url{https://huggingface.co/Yukang/LongAlpaca-13B} \\
LongChat-7B-v1.5 & 32k & LLaMA-2-7B & \url{https://huggingface.co/lmsys/longchat-7b-v1.5-32k} \\
Vicuna-7B-v1.5 & 16k* & LLaMA-2-7B & \url{https://huggingface.co/lmsys/vicuna-7b-v1.5-16k} \\
Vicuna-13B-v1.5 & 16k* & LLaMA-2-13B & \url{https://huggingface.co/lmsys/vicuna-13b-v1.5-16k} \\
LongAlign-7B & 64k & LLaMA-2-7B & \url{https://huggingface.co/THUDM/LongAlign-7B-64k} \\
LongAlign-13B & 64k & LLaMA-2-13B & \url{https://huggingface.co/THUDM/LongAlign-13B-64k} \\ 
LLaMA-3.1-8B & 128k & LLaMA-3.1-8B & \url{https://huggingface.co/meta-llama/Meta-Llama-3.1-8B-Instruct} \\ 
Phi-3-small & 128k & Phi-3-small & \url{https://huggingface.co/microsoft/Phi-3-small-128k-instruct} \\ \bottomrule

\end{tabular}
}
\end{table*}

In our experiments on ELITR-Bench, we compared responses generated by 12 LLMs with long-context capabilities. We included both commercial models and open long-context models in our benchmarking:
\begin{itemize}
    \item \textbf{GPT-3.5}, \textbf{GPT-4}~\citep{OpenAI2023} and \textbf{GPT-4o}\footnote{\url{https://openai.com/index/gpt-4o-system-card/}} are powerful commercial LLMs from OpenAI that have obtained state-of-the-art performance on a wide range of LLM benchmarks. We used the gpt-3.5-turbo-16k-0613, gpt-4-1106-preview, and gpt-4o-2024-05-13 checkpoints,\footnote{\url{https://platform.openai.com/docs/models/}} which enable a context length of 16k tokens for GPT-3.5 and 128k for GPT-4 and GPT-4o.
    \item \textbf{LongAlpaca-7B} and \textbf{LongAlpaca-13B} were obtained by fine-tuning LLaMA-2 models using the LongLoRA technique on the LongAlpaca dataset, both introduced in~\citet{chen2024longlora}. Their context size limit is 32k.
    \item \textbf{LongChat-7B-v1.5} is the LLaMa-2 version of the original LongChat-7B model~\citep{longchat2023}, trained on curated conversation data with rotary position embeddings~(RoPE)~\citep{Su2024rope}. It enables a context of at most 32k tokens.
    \item \textbf{Vicuna-7B-v1.5} and \textbf{Vicuna-13B-v1.5} were obtained by fine-tuning LLaMA-2 on the user-shared ShareGPT conversations, similarly to the original Vicuna model~\citep{vicuna2023}. Their context length is 16k~-- which we extrapolate to 32k at inference time using RoPE~\citep{Su2024rope}, to enable processing the longer meeting transcripts.
    \item \textbf{LongAlign-7B} and \textbf{LongAlign-13B} are based on the LongAlign recipe \citep{longalign} by fine-tuning LLaMA-2 models on synthetic long sequences using a compact batching strategy. Their maximum context size is 64k tokens.
    \item \textbf{LLaMA-3.1-8B} is the latest iteration (at the time of writing) of the LLaMA family of models from Meta AI~\citep{llama3}. This model enables a context limit of 128k due to its native long-context fine-tuning. We used the instruction-tuned version of the model in our experiments.
    \item \textbf{Phi-3-small} is the 3rd model of the Phi family of LLMs from Microsoft~\citep{phi3}. We adopted the `small' instruction-tuned version which has 7B parameters and accepts a context up to 128k tokens.
\end{itemize}

We summarize the details of the different long-context LLMs in Table~\ref{tab:compared-models}. We provide for each model its context size limit in tokens, its backbone model (i.e., the pre-trained model used for the fine-tuning), and the link to the model checkpoint on Huggingface for open models or the link to the relevant OpenAI documentation for proprietary models.

The inference was done on a single A100 GPU with 80GB memory. In preliminary experiments, we also attempted to include the Mistral-7B-Instruct-v0.2\footnote{\url{https://huggingface.co/mistralai/Mistral-7B-Instruct-v0.2}} model in our study, as this model supports a context of up to 32k tokens. However, running this model on ELITR-Bench led to a GPU out-of-memory error on the A100, and thus we discarded it.

\subsection{Configuration search and hyperparameter setting}
\label{app:hyperparam}

In our pilot experiments, we noted that the LLaMA-2-based open models retained for our study tended to be fairly impacted by the choice of the prompt and the inference hyperparameters. Therefore, we conducted a search on the inference configuration space to select appropriate hyperparameters for each LLaMA-2-based model.\footnote{In comparison to these, we found that GPT-3.5, GPT-4/4o, LLaMA-3.1-8B, and Phi-3-small were more robust to differences in the inference configuration. Therefore, we did not conduct a configuration search on these.} The configuration search was carried out in two steps on the dev set of ELITR-Bench-QA, in the single-turn mode. The evaluation was performed using GPT-4 as the evaluator, as described in the evaluation protocol in Section~\ref{sec:exp-setup}. 

In the first step of the search~-- whose results are given in Table~\ref{tab:step-1}~-- we varied three dimensions in the inference:
\begin{itemize}
    \item The decoding method, which was either greedy decoding or nucleus sampling with a temperature of 0.6 and top-p of 0.9;
    \item The use (or absence) of a chat template,\footnote{\url{https://huggingface.co/docs/transformers/main/en/chat_templating}} which modifies the prompt to integrate the same tags used in the fine-tuning stage~-- those tags varying across models;
    \item The use (or absence) of question-answer markers, which introduces to the prompt the tokens `QUESTION:' and `ANSWER:' before the question and the expected answer, respectively.
\end{itemize}
The specific chat template we adopted for each model is based on the one used during the model's fine-tuning: the LLaMA-2 template for LongAlpaca-7B and LongAlpaca-13B; the Vicuna template for LongChat-7B-v1.5, Vicuna-7B-v1.5 and Vicuna-13B-v1.5; and the LongAlign template for LongAlign-7B and LongAlign-13B.

In the second step of the search, we used the configuration that yielded the best results on the first step for each model and tested the impact of setting the repetition penalty hyperparameter to 1.1 (instead of the default 1.0 value) in the inference. The results of step 2 are provided in Table~\ref{tab:step-2}.

Ultimately, the following configurations were retained for each model:
\begin{itemize}
    \item \textbf{LongAlpaca-7B:} greedy decoding with a chat template and QA markers;
    \item \textbf{LongAlpaca-13B:} greedy decoding with a chat template;
    \item \textbf{LongChat-7B-v1.5:} greedy decoding with a chat template;
    \item \textbf{Vicuna-7B-v1.5:} nucleus sampling with QA markers;
    \item \textbf{Vicuna-13B-v1.5:} nucleus sampling with a chat template, QA markers, and repetition penalty;
    \item \textbf{LongAlign-7B:} greedy decoding with a chat template;
    \item \textbf{LongAlign-13B:} nucleus sampling with a chat template.
\end{itemize}
The cost of the two-step configuration search amounted to approximately \$150.\footnote{We assessed the cost of performing the evaluation of a single model on the 141 dev set questions to \$3 approximately. As we evaluated 7 models on 6 configurations in the first step, and 7 models on 1 configuration in the second step, this yields \$147.} To limit excessive expenses, we used the same model configuration for the different settings we experimented in (single-turn ELITR-Bench-QA, multi-turn ELITR-Bench-QA, and multi-turn ELITR-Bench-Conv).

For the proprietary models, \textbf{GPT-3.5}, \textbf{GPT-4} and \textbf{GPT-4o}, we used nucleus sampling (temperature = 0.6 and top-p = 0.9) with the standard OpenAI chat template. The \textbf{LLaMA-3.1-8B} and \textbf{Phi-3-small} models were run with greedy decoding and using their respective chat templates.

\begin{table*}[t]
\centering
\caption{Results of step 1 for our configuration search on ELITR-Bench-QA's dev set, in the single-turn mode. The configuration corresponding to using neither a chat template nor QA markers is not included as this was shown to severely underperform in our preliminary experiments.}
\label{tab:step-1}
\scalebox{0.9}{
\begin{tabular}{lll|rrrrrrr}
\toprule
\textbf{Decoding} & \begin{tabular}[c]{@{}l@{}}\textbf{Chat}\\ \textbf{templ.}\end{tabular} & \begin{tabular}[c]{@{}l@{}}\textbf{QA}\\ \textbf{mark.}\end{tabular} & \begin{tabular}[c]{@{}r@{}}\textbf{LongAl-}\\ \textbf{paca-7B}\end{tabular} & \begin{tabular}[c]{@{}r@{}}\textbf{LongAl-}\\ \textbf{paca-13B}\end{tabular} & \begin{tabular}[c]{@{}r@{}}\textbf{LongChat-}\\ \textbf{7B-v1.5}\end{tabular} & \begin{tabular}[c]{@{}r@{}}\textbf{Vicuna-}\\ \textbf{7B-v1.5}\end{tabular} & \begin{tabular}[c]{@{}r@{}}\textbf{Vicuna-}\\ \textbf{13B-v1.5}\end{tabular} & \begin{tabular}[c]{@{}r@{}}\textbf{LongAl-}\\ \textbf{ign-7B}\end{tabular} & \begin{tabular}[c]{@{}r@{}}\textbf{LongAl-}\\ \textbf{ign-13B}\end{tabular} \\ \midrule
Greedy & Y & Y & \textbf{5.89} & 6.13 & 6.22 & 4.94 & 5.19 & 6.04 & 6.16 \\
Greedy & Y & N & 5.55 & \textbf{6.17} & \textbf{6.60} & 5.38 & 5.13 & \textbf{6.11} & 6.16 \\
Greedy & N & Y & 5.89 & 5.87 & 6.23 & 5.05 & 4.71 & 5.43 & 5.94 \\
Nucleus & Y & Y & 5.18 & 5.91 & 6.19 & 4.99 & \textbf{5.70} & 5.67 & 6.25 \\
Nucleus & Y & N & 5.61 & 6.11 & 5.85 & 5.33 & 5.00 & 6.06 & \textbf{6.27} \\
Nucleus & N & Y & 5.58 & 5.96 & 5.91 & \textbf{5.42} & 4.89 & 5.18 & 5.99 \\ \bottomrule
\end{tabular}
}
\end{table*}

\begin{table*}[t]
\centering
\caption{Results of step 2 for our configuration search on ELITR-Bench-QA's dev set, in the single-turn mode. In the cases where we include a repetition penalty, we set the corresponding hyperparameter to 1.1 (instead of 1.0, the default value corresponding to no repetition penalty).}
\label{tab:step-2}
\scalebox{0.9}{
\begin{tabular}{l|rrrrrrr}
\toprule
\begin{tabular}[c]{@{}l@{}}\textbf{Repetition}\\ \textbf{penalty}\end{tabular} & \begin{tabular}[c]{@{}r@{}}\textbf{LongAl-}\\ \textbf{paca-7B}\end{tabular} & \begin{tabular}[c]{@{}r@{}}\textbf{LongAl-}\\ \textbf{paca-13B}\end{tabular} & \begin{tabular}[c]{@{}r@{}}\textbf{LongChat-}\\ \textbf{7B-v1.5}\end{tabular} & \begin{tabular}[c]{@{}r@{}}\textbf{Vicuna-}\\ \textbf{7B-v1.5}\end{tabular} & \begin{tabular}[c]{@{}r@{}}\textbf{Vicuna-}\\ \textbf{13B-v1.5}\end{tabular} & \begin{tabular}[c]{@{}r@{}}\textbf{LongAl-}\\ \textbf{ign-7B}\end{tabular} & \begin{tabular}[c]{@{}r@{}}\textbf{LongAl-}\\ \textbf{ign-13B}\end{tabular} \\ \midrule
Y & 5.80 & 5.73 & 6.11 & 4.90 & \textbf{5.92} & 5.90 & 6.21 \\
N & \textbf{5.89} & \textbf{6.17} & \textbf{6.60} & \textbf{5.42} & 5.70 & \textbf{6.11} & \textbf{6.27} \\ \bottomrule
\end{tabular}
}
\end{table*}

\section{Additional experimental results}

\subsection{Variance over seeded run results}
\label{app:test-variance}

To account for the seed-dependent variability in the evaluation, we performed 3 seeded runs on the test set. The set of seeds used is \{2023, 2024, 2025\}. The results are reported in Table~\ref{tab:seeded-runs}, where we indicate for each (model, setting) pair the mean score over the 3 seeds as well as the sample standard deviation. 

Note that the same seed is used both for the response generation part and the GPT-4-based evaluation part, as both can be sources of variance in the reported results. Based on our configuration search (see Appendix~\ref{app:hyperparam}), some of the response generation models were set to use greedy decoding: LongAlpaca-7B, LongAlpaca-13B, LongChat-7B-v1.5, LongAlign-7B, LLaMA-3.1-8B and Phi-3-small. For these models, the response generation is deterministic and the only source of variance is that of the GPT-4 evaluator.

The results from Table~\ref{tab:seeded-runs} show that the variance across settings is fairly different. In the single-turn ELITR-Bench-QA setting, the standard deviation for all models remain relatively low, even for the models that use nucleus sampling (GPT-3.5, GPT-4, GPT-4o, Vicuna-7B-v1.5, Vicuna-13B-v1.5, LongAlign-13B). However, in the multi-turn settings, we observe an increased standard deviation for those same models overall, in particular for Vicuna-7B-v1.5 and LongAlign-13B. We hypothesize that the sequence of questions asked in the same conversation in the multi-turn setting causes different seeded runs to cumulate errors and slightly diverge along the course of the conversation.

\begin{table*}[t]
\centering
\caption{Results for the seeded runs on the test set for different ELITR-Bench settings. The reported numbers correspond to the mean score ± sample standard deviation computed over 3 seeds. Boldface numbers correspond to the best performance among proprietary or open models. The results for GPT-3.5 are omitted in the multi-turn setting as the context length exceeded the 16k limit of this model.}
\label{tab:seeded-runs}
\scalebox{0.9}{
\begin{tabular}{lrrr}
\toprule
\multirow{2}{*}[-0.25cm]{\textbf{Model}} & \multicolumn{1}{c}{\textbf{Single-turn}} & \multicolumn{2}{c}{\textbf{Multi-turn}} \\ \cmidrule(l{4pt}r{4pt}){2-2} \cmidrule(l{4pt}r{4pt}){3-4} 
 & \textbf{\begin{tabular}[c]{@{}r@{}}ELITR-Bench-QA\\ (test set)\end{tabular}} & \textbf{\begin{tabular}[c]{@{}r@{}}ELITR-Bench-QA\\ (test set)\end{tabular}} & \textbf{\begin{tabular}[c]{@{}r@{}}ELITR-Bench-Conv\\ (test set)\end{tabular}} \\ \midrule
GPT-3.5 & 7.44 ± 0.12 & - & - \\
GPT-4 & 8.38 ± 0.07 & \textbf{8.42 ± 0.09} & 8.36 ± 0.12 \\
GPT-4o & \textbf{8.44 ± 0.04} & 8.38 ± 0.04 & \textbf{8.41 ± 0.10} \\ \midrule
LongAlpaca-7B & 5.60 ± 0.06 & 4.84 ± 0.02 & 4.58 ± 0.04 \\
LongAlpaca-13B & 6.25 ± 0.05 & 4.71 ± 0.01 & 4.74 ± 0.06 \\
LongChat-7B-v1.5 & 5.78 ± 0.06 & 4.17 ± 0.07 & 4.31 ± 0.07 \\
Vicuna-7B-v1.5 & 5.61 ± 0.17 & 4.61 ± 0.26 & 4.69 ± 0.34 \\
Vicuna-13B-v1.5 & 6.52 ± 0.16 & 5.67 ± 0.10 & 5.78 ± 0.13 \\
LongAlign-7B & 6.46 ± 0.07 & 4.47 ± 0.01 & 5.06 ± 0.03 \\
LongAlign-13B & 6.33 ± 0.09 & 5.33 ± 0.47 & 4.95 ± 0.22 \\
LLaMA-3.1-8B & \textbf{7.83 ± 0.03} & \textbf{7.81 ± 0.02} & \textbf{7.78 ± 0.03} \\
Phi-3-small & 7.34 ± 0.05 & 7.52 ± 0.08 & 7.38 ± 0.07 \\ \bottomrule
\end{tabular}
}
\end{table*}

\subsection{Results on QA/Conv differentiating questions}
\label{sec:differentiating-questions}

\begin{figure}[t]
\centering
\includegraphics[width = 0.95\textwidth]{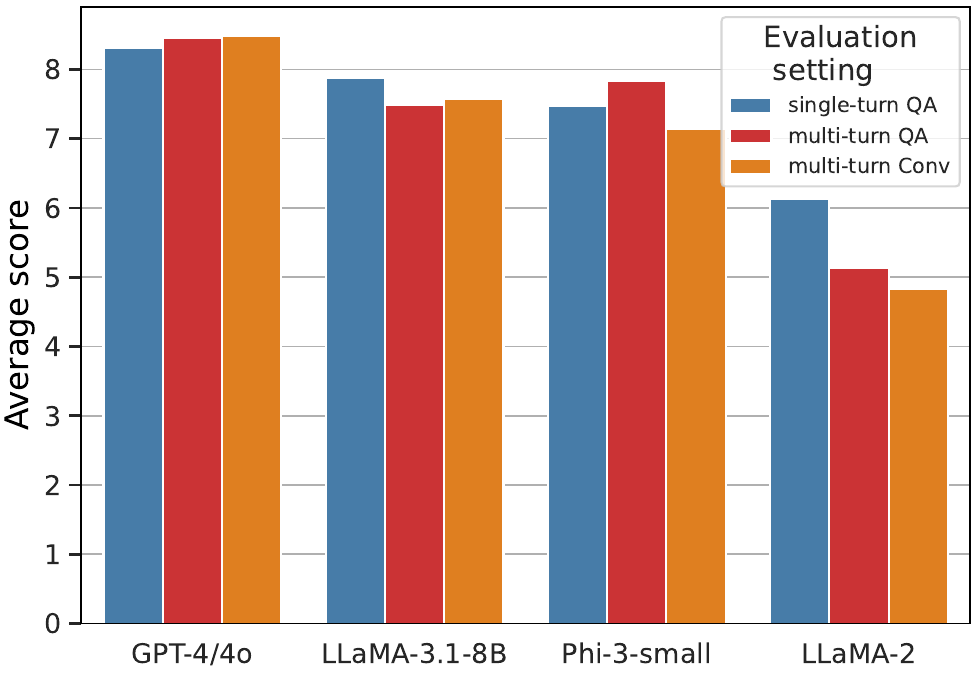} 
\caption{Results restricted to QA/Conv differentiating questions. The score reported for each model and evaluation setting corresponds to the average of the scores obtained on the dev subset (16 questions) and the test subset (17 questions). Best viewed in color.}
\label{fig:differentiating-questions}       
\end{figure}

As introduced in Section~\ref{sec:elitr-bench}, some of the questions differ between ELITR-Bench-QA and ELITR-Bench-Conv and typically contain pronominal references or ellipses in the Conv setting, which makes them particularly challenging to tackle. In this section, we look at the results on this subset of differentiating questions~-- both in their QA and Conv versions~-- and study the impact of using the single-turn or multi-turn mode. The results are provided in Fig.~\ref{fig:differentiating-questions}, which compares 3 settings: single-turn mode with QA questions, multi-turn mode with QA questions, and multi-turn mode with Conv questions. The reported scores are averaged over the dev and test sets' differentiating questions (respectively, 16 and 17 questions) to make up for the limited size of these subsets.

Similarly with what we observed in Table~\ref{tab:main-results}, we notice again a clear difference between the top-performing models~-- GPT-4/4o, LLaMA-3.1-8B, and Phi-3-small~-- and the LLaMA-2-based models: the performance of the former is maintained from single-turn to multi-turn, whereas the performance of the latter notably degrades. In contrast with our previous findings that showed little to no difference between the results on ELITR-Bench-QA and ELITR-Bench-Conv for the multi-turn mode, we observe this time that the average score decreases from QA to Conv for LLaMA-2-based models. While the difference is small, this trend is expected as the Conv questions in this subset are more challenging to answer due to the pronominal references. We hypothesize that the opposite trends identified for top-performing models and LLaMA-2-based models might be explained by a `snowballing' effect that causes an error propagation in lower-performing models. It is also possible that GPT-4/4o and the more recent open LLMs (LLaMA-3.1-8B and Phi-3) have been fine-tuned on a larger volume of conversational (i.e., generally multi-turn) data which could imply a greater adaptation to this setting.

\subsection{Investigation of a ``lost in the middle'' effect} 
\label{app:lost-in-the-middle}

\begin{table}[t]
\centering
\caption{Results of a one-tailed Welch's t-test on the alternative hypothesis ``The average score for questions with middle-position answers is lower than the average score of other questions'', to verify the presence or absence of a ``lost in the middle'' effect \citep{liu2023lost}. Boldface numbers denote statistically significant results (p-value $<$ 0.05).}
\label{tab:lost-middle}
\begin{tabular}{lr}
\toprule
\textbf{Model} & \textbf{p-value} \\ \midrule
GPT-3.5 & 0.466 \\
GPT-4 & 0.372 \\
GPT-4o & 0.754 \\ \midrule
LongAlpaca-7B & 0.713 \\
LongAlpaca-13B & 0.265 \\
LongChat-7B-v1.5 & \textbf{0.032} \\
Vicuna-7B-v1.5 & \textbf{0.046} \\
Vicuna-13B-v1.5 & 0.469 \\
LongAlign-7B & 0.409 \\
LongAlign-13B & 0.413 \\
LLaMA-3.1-8B & 0.308 \\
Phi-3-small & 0.541 \\ \bottomrule
\end{tabular}
\end{table}

Past work reported a ``lost in the middle'' effect~\citep{liu2023lost}, stating that the middle of a model's context tends to be overlooked more often than the beginning or end of the context. To further investigate this phenomenon in our dataset, we conducted a statistical hypothesis test on the scores obtained by each individual model. Specifically, we ran a one-tailed Welch's t-test~\citep{Welch1947} with the following alternative hypothesis: ``The average score for questions with middle-position answers is lower than the average score of other questions.'' The p-values obtained for each model's set of scores are given in Table~\ref{tab:lost-middle}. Interestingly, we observe that the ``lost in the middle'' hypothesis is statistically verified (p-value $<$ 0.05) for only two models: LongChat-7B-v1.5 (p-value $=$ 0.032) and Vicuna-7B-v1.5 (p-value $=$ 0.046). While we do not have a clear explanation about which of these two models' characteristics caused that effect, these models have in common that they are based on LLaMA-2-7B and were trained by the same LMSYS organization. It is then possible~-- although purely hypothetical~-- that the specific fine-tuning recipe followed by LMSYS on LLaMA-2-7B for these two models led to the ``lost in the middle'' effect.

\section{Crowdsourcing study details}
\label{app:prolific}

\begin{figure*}[t]
\centering
\includegraphics[width = 0.7\textwidth]{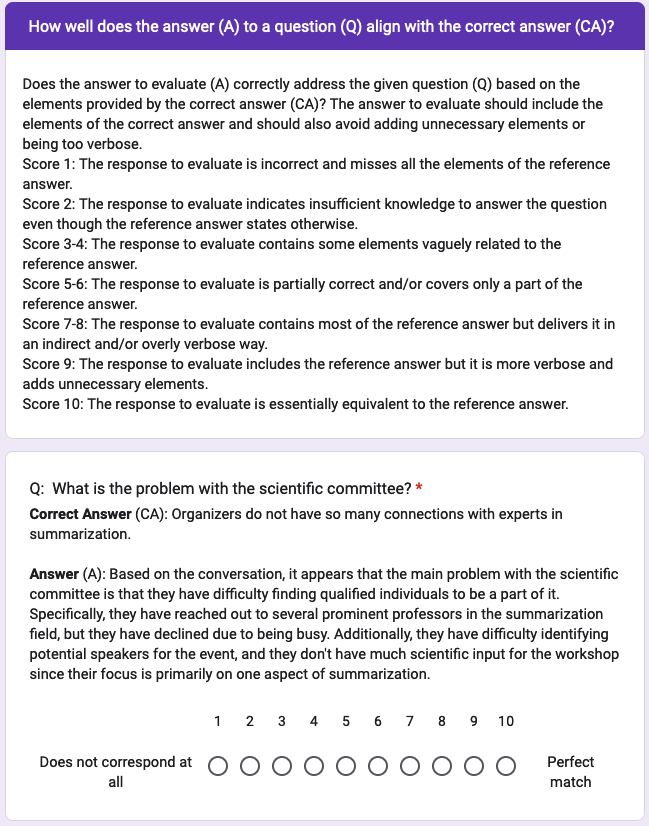} 
\caption{Interface for our Prolific crowdsourcing study to collect Silver Human score annotations.}
\label{fig:prolific-interface}       
\end{figure*}

Our \textit{Silver Human} evaluation is based on a crowdsourcing study using the Prolific\footnote{\url{https://www.prolific.com/}} platform. A task in this study consists in scoring the responses of the 3 considered models (GPT-4, Vicuna-13B-v1.5, and LongAlpaca-7B) for all the questions of a single meeting~-- out of 8 meetings in the test set. For each meeting, we hired 10 annotators, without constraining the 10 annotators to be the same across meetings. 
Participants were screened based on their primary language (English) and domain expertise (including Computer Science, Information Technology, Engineering, or Mathematics). Each participant received £9 per hour when completing a task (with each task comprising approximately 40-50 questions for assessment). We estimated the task duration to be around 30 minutes~-- our post-analysis indicated a median time spent per study ranging between 16 and 29 minutes depending on the meeting.
We discarded the annotations that were flagged as too inconsistent with Gold Human scores, and hired new annotators when needed until we had a satisfactory set of 10 annotators per meeting.  In total, the crowdsourced Prolific evaluation cost was £400.

The guidelines provided for this study start with general information about the task as well as the 10-point score rubric given in Fig.~\ref{fig:score-rubric-gpt4}, in order to help annotators calibrate their scores with concrete criteria. Then the interface presents a tuple composed of a question, its ground-truth answer, and an LLM response to evaluate. Given this tuple, the annotator is asked to grade the LLM response with a score ranging from 1 to 10, following the provided score rubric. A screenshot of our interface is shown in Fig.~\ref{fig:prolific-interface}.

To measure the inter-annotator agreement, we used the intra-class correlation (ICC) coefficient \citep{koo2016guideline} which assesses how consistent annotators' scores are for every (question, ground-truth answer, LLM response) tuple. The ICC results are detailed in Table~\ref{tab:icc} for each individual meeting and overall. For individual meetings, we report the two-way coefficient ICC(2,k) as the set of hired annotators is the same across all the questions of a given meeting. For the result over all meetings, we used instead the one-way coefficient ICC(1,k) since the set of annotators differs across meetings. Most of the ICC coefficients being above 0.9 suggests an excellent inter-annotator agreement, following the interpretation guidelines from~\citep{koo2016guideline}.

\begin{table}[t]
\centering
\caption{Intra-class correlation (ICC) coefficients across annotators from the Prolific crowdsourcing study, corresponding to the Silver Human evaluator.}
\label{tab:icc}
\begin{tabular}{cr}
\toprule
\textbf{Meeting ID} & \textbf{ICC} \\ \midrule
01 & 0.872 \\
02 & 0.964 \\
03 & 0.912 \\
04 & 0.941 \\
05 & 0.906 \\
06 & 0.940 \\
07 & 0.936 \\
08 & 0.942 \\ \midrule
all & 0.965 \\ \bottomrule
\end{tabular}
\end{table}

\section{ELITR-Bench excerpt}
\label{app:excerpt}

\begin{table*}[t]
\small
\caption{Small excerpt of meeting 010 from ELITR's dev set, with sample questions and answers related to the same meeting from ELITR-Bench.}
\label{tab:elitr-example}
\begin{tabular}{|l|p{0.7\linewidth}|}
\hline
\textbf{Transcript excerpt} & \begin{tabular}{@{}p{\linewidth}@{}}
... \\
(PERSON19) Just \textless unintelligible/\textgreater{} like a virtual machine image. \\
(PERSON10) Yeah, yeah. \\
(PERSON19) You just fire up, an- anyone can fire up, it's not like you have to you have to call- \\
(PERSON10) Yeah. \\
(PERSON19) Like [ORGANIZATION11], get them to run it. \\
(PERSON10) Yeah. \\
(PERSON19) I I don't know that's easier, but I mean it it's more more flexible. \\
(PERSON10) Yeah, yeah. \\
I haven't since I haven't really done it, it's uh, it's hard for me to access, so we- \\
(PERSON19) I know, I know. \\
(PERSON10) You know. \\
Uh, okay, so that's good, we know what to do. I don't know whether we'll manage to have these systems package before the demo, but hopefully uh, there won't be any power outage an our uh, at our site. \\
(PERSON19) \textless laugh/\textgreater{} \\
(PERSON10) \textless laugh/\textgreater{} \\
So that was the 1 thing, that I've learnt, that we must not uh, that that we must have uh, rep- replicated uh, components across the site. \\
...
\end{tabular} \\ \hline
\hline
\textbf{Question (What)} & Which risk, related to the demo, was discussed? \\ \hline
\textbf{Answer} & Power outages at [ORGANIZATION2] \\ \hline
\textbf{Question (Who)} & Which entity is running the translation module? \\ \hline
\textbf{Answer} & [ORGANIZATION11] \\ \hline
\textbf{Question (What)} & What should be frozen 1 or 2 weeks before the demo? \\ \hline
\textbf{Answer} & The stable components of the systems should be frozen 1-2 weeks before the demo \\ \hline
\textbf{Question (When)} & When should the recorded demo be provided? \\ \hline
\textbf{Answer} & 17th of June \\ \hline
\end{tabular}
\end{table*}

We provide in Table~\ref{tab:elitr-example} an excerpt of meeting 010 from the dev set of the original ELITR corpus~\citep{nedoluzhko-etal-2022-elitr}. Entities, such as (PERSON10), (PERSON19), and [ORGANIZATION11], have been de-identified in the original work for the sake of anonymization. Below the excerpt, we provide 4 questions (and their respective answers) related to the same meeting, which have been added through the proposed ELITR-Bench. For each question, we indicate its type between brackets (i.e., \textit{Who}, \textit{What}, \textit{When}, or \textit{How many}).

\section{Prompts}
\label{app:prompt}

\begin{figure*}[t]
  \centering
  \begin{tabular}{|c|}
    \hline\\
    \parbox{12cm}{%
        The following is the transcript of a meeting with multiple participants, where utterances start with the speaker's anonymized name (for instance (PERSON4)) and may span over several lines.\\\\
        \textcolor{blue}{\{transcript\}}\\\\
        As a professional conversational assistant, your task is to answer questions about the meeting by making inferences from the provided transcript.\\
    } \\
    \hline
  \end{tabular}
  \caption{Answer prompt used to obtain LLMs' responses. Questions are appended to this prompt as described in Section~\ref{sec:exp-setup}. The element in blue and enclosed in curly brackets corresponds to a meeting-specific text span that is dynamically adapted.}
  \label{fig:answer-prompt}
\end{figure*}

In this section, we list the different prompts used in the paper, both for response generation and evaluation. The prompt for response generation follows the same general template given in Fig.~\ref{fig:answer-prompt} for every evaluated model~-- both proprietary and open models. Then, questions and answers are appended to the prompt as described in Section~\ref{sec:exp-setup}~-- either a single question per conversation in the single-turn mode, or all the questions of a meeting in sequence in the multi-turn mode. As detailed in Appendix~\ref{app:hyperparam}, we slightly modify this base prompt depending on the model-specific selected configuration. As a reminder, these alterations may take two forms: the use of a chat template (which only adds special tags to the prompt) and the use of question-answer markers (which add `QUESTION:' before a question and `ANSWER:' before an answer).

The prompts that we used for evaluation are inspired from the prompt originally proposed in~\citep{kim2023prometheus} and include: the question, the response to evaluate, the ground-truth answer, and a score rubric. Note that the transcript is not included in the evaluation prompt as the question and ground-truth answer should provide sufficient information to assess the correctness of the response to evaluate. The full prompts are given in Fig.~\ref{fig:eval-prompt-gpt4} for the GPT-4 evaluator and in Fig.~\ref{fig:eval-prompt-prometheus} for the Prometheus evaluator. Their score rubrics are shown in Fig.~\ref{fig:score-rubric-gpt4} and Fig.~\ref{fig:score-rubric-prometheus}, respectively. For Prometheus, we had to adapt the 10-point score scale to a 5-point scale to match the format used when this model was fine-tuned~\citep{kim2023prometheus}. The 5-point rubric was defined to retain the main criteria expressed in the 10-point rubric and minimally alter it to enable a fair comparison between the two evaluators.

\begin{figure*}[t]
  \centering
  \begin{tabular}{|c|}
    \hline\\
    \parbox{13cm}{%
        \#\#\# Task description:\\
        You are provided below with a question, a response to evaluate, a reference answer that gets the maximum score of 10, and a score rubric representing evaluation criteria.\\
        1. Write a detailed feedback that assess the quality of the response strictly based on the given score rubric, not evaluating in general.\\
        2. After writing a feedback, write a score that is an integer between 1 and 10. You should refer to the score rubric.\\
        3. The output format should first include the feedback and then indicate the integer score in \textbackslash boxed\{\}.\\
        4. Please do not generate any other opening, closing, and explanations.\\\\
        \#\#\# Question:\\
        \textcolor{blue}{\{question\}}\\\\
        \#\#\# Response to evaluate:\\
        \textcolor{blue}{\{response\}}\\\\
        \#\#\# Reference answer (score 10):\\
        \textcolor{blue}{\{reference\}}\\\\
        \#\#\# Score rubric:\\
        \textcolor{blue}{\{rubric\}}\\\\
        \#\#\# Feedback:
        \\
    } \\
    \hline
  \end{tabular}
  \caption{Evaluation prompt for the GPT-4 evaluator, inspired from~\citet{kim2023prometheus}. The elements in blue and enclosed in curly brackets correspond to question-specific text spans that are dynamically adapted.}
  \label{fig:eval-prompt-gpt4}
\end{figure*}

\begin{figure*}[t]
  \centering
  \begin{tabular}{|c|}
    \hline\\
    \parbox{13cm}{%
        {[Does the response to evaluate correctly address the given question based on the elements provided by the reference answer? The response should include the elements of the reference answer and should also avoid adding unnecessary elements or being too verbose.]}\\
        \textbf{Score 1:} The response to evaluate is incorrect and misses all the elements of the reference answer.\\
        \textbf{Score 2:} The response to evaluate indicates insufficient knowledge to answer the question even though the reference answer states otherwise.\\
        \textbf{Score 3-4:} The response to evaluate contains some elements vaguely related to the reference answer.\\
        \textbf{Score 5-6:} The response to evaluate is partially correct and/or covers only a part of the reference answer.\\
        \textbf{Score 7-8:} The response to evaluate contains most of the reference answer but delivers it in an indirect and/or overly verbose way.\\
        \textbf{Score 9:} The response to evaluate includes the reference answer but it is more verbose and adds unnecessary elements.\\
        \textbf{Score 10:} The response to evaluate is essentially equivalent to the reference answer.
        \\
    } \\
    \hline
  \end{tabular}
  \caption{Score rubric for the GPT-4 evaluator. Boldface is added for the sake of readability and is not included in the actual prompt.}
  \label{fig:score-rubric-gpt4}
\end{figure*}

\begin{figure*}[t]
  \centering
  \begin{tabular}{|c|}
    \hline\\
    \parbox{13cm}{%
        \#\#\# Task description:\\
        You are provided below with a question, a response to evaluate, a reference answer that gets the maximum score of 5, and a score rubric representing evaluation criteria.\\
        1. Write a detailed feedback that assesses the quality of the response strictly based on the given score rubric, not evaluating in general.\\
        2. After writing a feedback, write a score that is an integer between 1 and 5. You should refer to the score rubric.\\
        3. The output format should look as follows: "Feedback: (write the quality assessment feedback) [RESULT] (an integer number between 1 and 5)".\\
        4. Please do not generate any other opening, closing, and explanations.\\\\
        
        \#\#\# Question:\\
        \textcolor{blue}{\{question\}}\\\\
        
        \#\#\# Response to evaluate:\\
        \textcolor{blue}{\{response\}}\\\\
        
        \#\#\# Reference answer (score 5):\\
        \textcolor{blue}{\{reference\}}\\\\
        
        \#\#\# Score rubric:\\
        \textcolor{blue}{\{rubric\}}\\\\
        
        \#\#\# Feedback:
        \\
    } \\
    \hline
  \end{tabular}
  \caption{Evaluation prompt for the Prometheus evaluator, inspired from~\citet{kim2023prometheus}. The elements in blue and enclosed in curly brackets correspond to question-specific text spans that are dynamically adapted.}
  \label{fig:eval-prompt-prometheus}
\end{figure*}

\begin{figure*}[t]
  \centering
  \begin{tabular}{|c|}
    \hline\\
    \parbox{13cm}{%
        {[Does the response to evaluate correctly address the given question based on the elements provided by the reference answer? The response should include the elements of the reference answer and should also avoid adding unnecessary elements or being too verbose.]}\\
        \textbf{Score 1:} The response to evaluate is incorrect and misses all the elements of the reference answer.\\
        \textbf{Score 2:} The response to evaluate contains some elements vaguely related to the reference answer.\\
        \textbf{Score 3:} The response to evaluate is partially correct and/or covers only a part of the reference answer.\\
        \textbf{Score 4:} The response to evaluate contains most of the reference answer but delivers it in an indirect and/or overly verbose way.\\
        \textbf{Score 5:} The response to evaluate is essentially equivalent to the reference answer.
        \\
    } \\
    \hline
  \end{tabular}
  \caption{Score rubric for the Prometheus evaluator. Boldface is added for the sake of readability and is not included in the actual prompt.}
  \label{fig:score-rubric-prometheus}
\end{figure*}

\end{document}